\documentclass[letterpaper]{article} 
\usepackage{aaai2027}  
\nocopyright  
\usepackage[hyphens]{url}
\usepackage{graphicx}
\usepackage{natbib}
\usepackage{caption}
\usepackage{amsmath,amssymb}
\usepackage{amsthm}
\usepackage{booktabs}
\usepackage{array}
\usepackage{algorithm}
\usepackage{algorithmic}

\newcommand{\detn}{d}
\newcommand{\obj}{o}
\newcommand{\bbi}{b}
\newcommand{\bbo}{B}
\newcommand{\cfeat}{f}
\newcommand{\ofeat}{g}
\newcommand{\lbl}{\ell}
\newcommand{\lblstar}{\lbl^{\!*}}
\newcommand{\Vol}{\mathrm{Vol}}
\newcommand{\Ind}{\mathbf{1}}

\newcommand{\aggm}{\tilde{\varphi}}
\newcommand{\method}{OVIP-SG}

\title{\method{}: Open-Vocabulary Instance-Preserving Scene Graphs for Mapping and Retrieval of Small, Fine-Grained Objects}
\author{
    Tianjing Hao\textsuperscript{\rm 1,*\textdagger},
    Haiyu Lan\textsuperscript{\rm 2,*\textdagger},
    Angsong Li\textsuperscript{\rm 2},
    Cheng Chen\textsuperscript{\rm 2},
    Enyu Li\textsuperscript{\rm 2},\\
    Jiarui Yang\textsuperscript{\rm 2},
    Yuning Su\textsuperscript{\rm 2},
    Peiwen Lin\textsuperscript{\rm 2},
    Wang Chuang\textsuperscript{\rm 2}
}
\affiliations{
    \textsuperscript{\rm 1}Xi'an Jiaotong University\\
    \textsuperscript{\rm 2}AgiBot
}

\begin{document}
\maketitle
\begingroup
  \renewcommand{\thefootnote}{\fnsymbol{footnote}}%
  \footnotetext[1]{These authors contributed equally. Tianjing Hao's work was
    completed during an internship at AgiBot.}%
  \footnotetext[2]{Corresponding authors: Tianjing Hao and Haiyu Lan.}%
\endgroup

\begin{abstract}
Integrating open-vocabulary perception into object-level 3D scene graphs is a
double-edged sword. While vision-language detectors recover long-tail categories and
small, fine-grained objects overlooked by closed-set models, they also tend to
fragment large surfaces and merge small objects into larger neighboring objects,
compromising instance-level consistency and undermining mapping fidelity. Moreover,
existing methods struggle to retrieve previously unmapped targets or determine whether
a queried object is absent, hindering robust embodied open-world navigation and
exploration. We present \method{}, a unified framework for instance-preserving semantic
mapping, functional scene partitioning, and language-guided small, fine-grained object
retrieval. \method{} uses a vision-language model (VLM) to enumerate scene-specific
categories for robust open-world detection. Symmetric 3D Intersection over Union (IoU)
association and area-weighted feature fusion preserve small independent instances,
while VLM-inferred object functions partition scenes into compact functional search
regions. A four-stage cascaded retrieval pipeline further incorporates voxel voting and
determines target absence from exploration coverage. Under a unified evaluation protocol
on Replica, \method{} outperforms ConceptGraphs by 6.31 points in class-mean
accuracy (mAcc) and 5.15 points in frequency-weighted mIoU (F-mIoU) while achieving a
class-agnostic native-instance Panoptic Quality (PQ) of 0.398. It reduces the search area
to 21.8\% of the indoor floor space and reaches 0.773 balanced accuracy for
object-presence classification. Real-world robotic experiments further demonstrate its
practical effectiveness.
Code is available at \textcolor{blue}{\url{https://github.com/Agibot-Spatial-Intelligence/OVIP-SG}}.
\end{abstract}

\section{Introduction}
\label{sec:intro}

Object-level open-vocabulary maps have made semantic representations more useful for
embodied agents. However, existing systems, including
ConceptGraphs~\citep{Gu2024Conceptgraphs},
ConceptFusion~\citep{Jatavallabhula2023Conceptfusion}, and
HOV-SG~\citep{Werby2024Hierarchical}, exhibit limited instance fidelity and retrieval
reliability on real robots. Small objects
may be absorbed into furniture or fragmented, while large surfaces can be mislabeled.
Once small objects are lost during map construction, they cannot be located or
retrieved later.

\begin{figure}[t]
\centering
\includegraphics[width=\columnwidth]{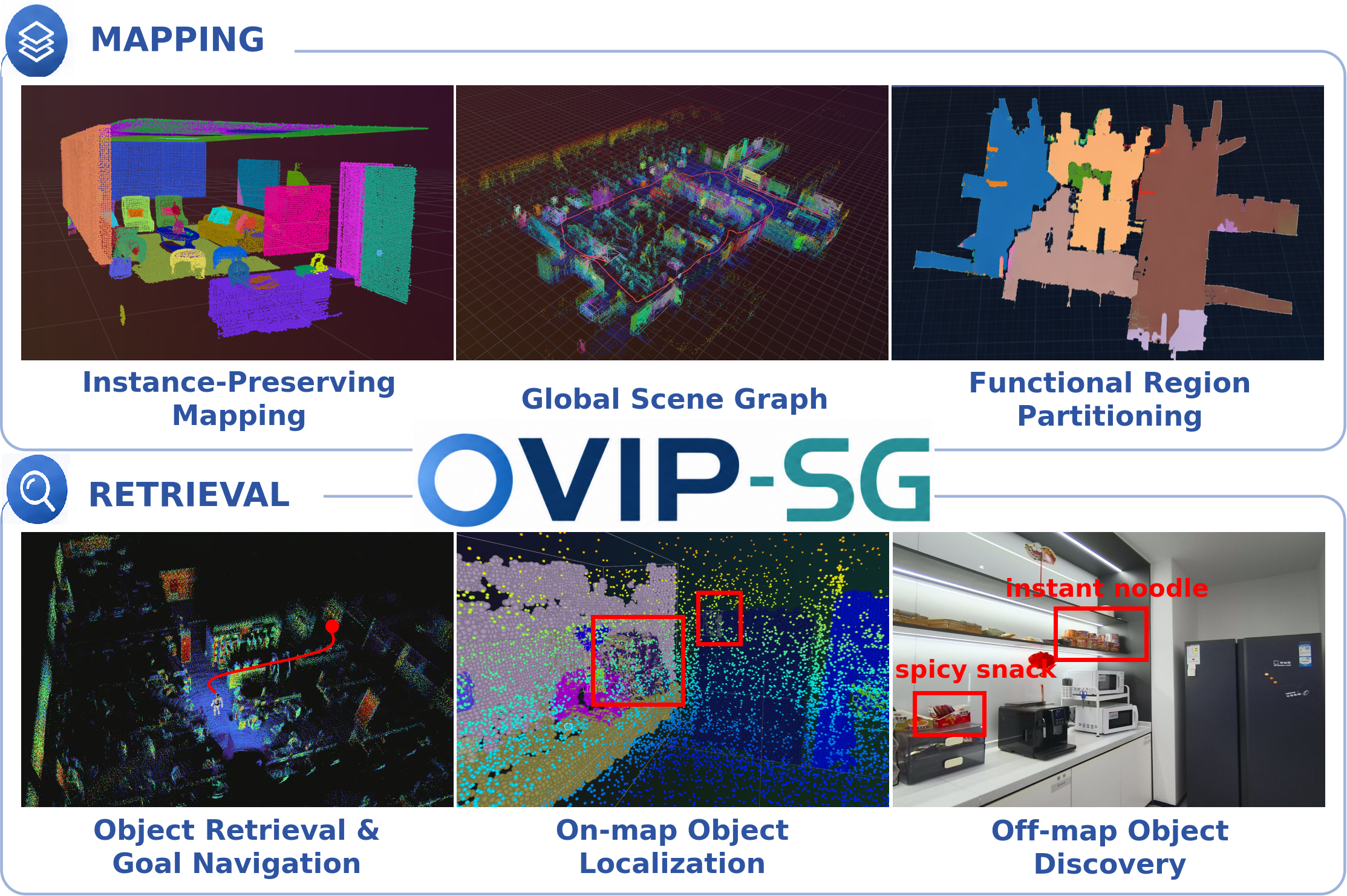}
\caption{Overview of the two capabilities in \method{}. \textbf{Mapping} (top),
from left to right: instance-preserving mapping, the global scene graph, and
functional-region partitioning. \textbf{Retrieval} (bottom), from left to
right: goal-directed object retrieval, on-map object localization, and cascaded
active retrieval for off-map object discovery.}
\label{fig:teaser}
\end{figure}

We target two capabilities: \textbf{Task~A}, \emph{high-fidelity
semantic mapping} that keeps small objects as their own instances and labels large
ones correctly; and \textbf{Task~B}, \emph{small, fine-grained object retrieval and
navigation} that grounds a free-form query to a 3D goal and path even for unmapped
targets---actively exploring when 3D information is missing and returning a
coverage-conditioned negative decision when search finds no target. Mapping is
evaluated on Replica; navigation is evaluated on ReplicaCAD and HM3D scans in
simulation. Online, closed-loop deployments in office and hotel environments provide
qualitative real-robot evidence. Fig.~\ref{fig:teaser} summarizes the mapping and
retrieval capabilities addressed in this work.

Task-A fidelity requires fixing two failure modes: \emph{asymmetric containment}
merges a small object wholesale into enclosing furniture, and count-averaged box-crop
CLIP fusion lets frequent small detections dominate a large object's feature.
\method{} uses an \emph{Enumerate-to-Ground} (E2G) front-end with a \emph{Frame-Adaptive Vocabulary Gate} (FAV-Gate) to obtain open-vocabulary detections. Its \emph{Instance-Preserving Association} (IPA) combines \emph{Instance-Preserving Symmetric-IoU Merge} (IPSM) and \emph{Label-Aware Merge Penalty} (LAMP) to protect small instances; \emph{Area-Weighted Fusion} (AWF) then updates each associated instance's feature. The VLM is called only when fine-grained semantic reasoning is required; routine cases are resolved from on-device geometric and CLIP evidence.

\noindent\textbf{Contributions.} Over prior open-vocabulary maps we offer
three capability gains, each via a concrete mechanism:
\begin{itemize}
  \item \textbf{(A) Higher-fidelity mapping.} IPA (IPSM + LAMP) and AWF operate on
  the E2G front-end with FAV-Gate (\S\ref{sec:e2g}--\S\ref{sec:ipa}).
  \item \textbf{(B) Small, fine-grained object retrieval and navigation.}
  \emph{Cascaded Active Retrieval with Absence Assessment} (CARA): a four-tier cascade whose
  \emph{Voxel-Vote Confirmation} (VVC) engine localizes off-map targets and returns a
  coverage-conditioned $\varnothing$-decision after exhaustive search (\S\ref{sec:cara}).
  \item \textbf{(C) Semantic room segmentation.} \emph{Knowledge-Guided Semantic
  Watershed} (KGSW) partitions rooms by VLM-defined function rather than wall
  geometry, giving (B) compact region priors.
\end{itemize}

\noindent On Replica, under the per-scene protocol shared by all baselines, \method{}
improves both class-mean accuracy ($+6.31$ mAcc) and frequency-weighted F-mIoU
($+5.15$) over the ConceptGraphs baseline and raises class-agnostic native-instance
PQ from $0.338$ to $0.398$.

\section{Related Work}
\label{sec:related}
We organize prior work around open-vocabulary mapping, scene organization, and
fine-grained retrieval and navigation.

\begin{figure*}[!t]
\centering
\includegraphics[width=\linewidth]{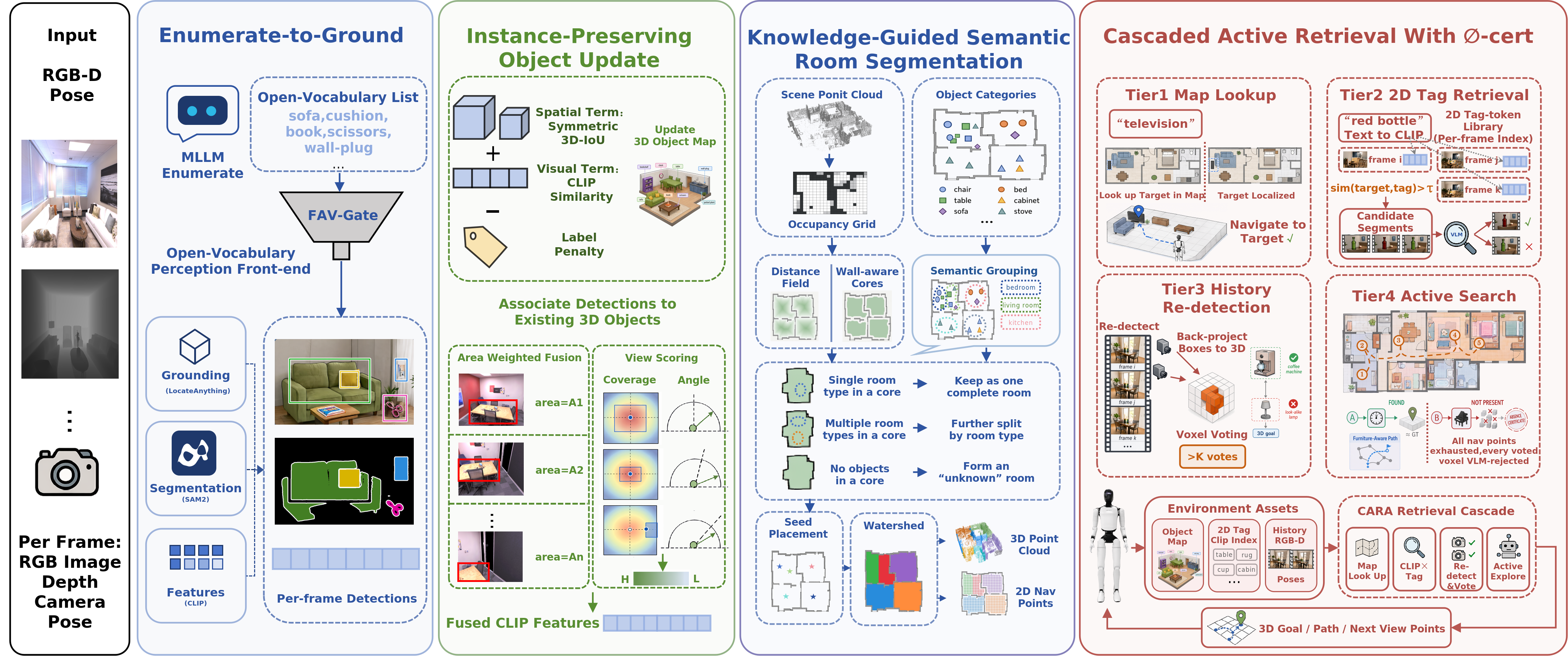}
\caption{Overview of \method{}. E2G with FAV-Gate produces open-vocabulary detections;
instance-preserving association (IPA) and view-area feature fusion build a
high-fidelity object map (Task A). KGSW organizes the map into functional regions,
while CARA performs small, fine-grained object retrieval and navigation through map lookup,
2D-tag retrieval, history re-detection, and active search with coverage-conditioned
negative decisions (Task B).}
\label{fig:overview}
\end{figure*}

\subsection{Open-vocabulary 3D mapping and semantic segmentation}
\label{sec:rw-mapping}
Closed-set systems fuse fixed labels into 3D~\citep{McCormac2017SemanticFusion,Rosinol2020Kimera,Grinvald2019Volumetric};
open-vocabulary work projects CLIP features into points, fields, and queryable maps
for 3D segmentation or retrieval~\citep{Peng2023OpenScene,Kerr2023LERF,Shafiullah2023CLIPFields,Huang2023Visual,Chen2023Openvocabulary,Yamazaki2024OpenFusion,Takmaz2023OpenMask3D,Lu2023OVIR3D,Ding2023PLA}.
ConceptFusion~\citep{Jatavallabhula2023Conceptfusion} and
ConceptGraphs~\citep{Gu2024Conceptgraphs} still merge small objects or bias features;
E2G/FAV-Gate, IPA, and AWF target these errors.

\subsection{3D scene graphs and room segmentation}
\label{sec:rw-graphs}
Scene graphs span object-relation, incremental RGB-D, metric-semantic, and
open-vocabulary hierarchies~\citep{Armeni20193D,Wu2021SceneGraphFusion,Hughes2022Hydra,Werby2024Hierarchical,Maggio2024Iclioi}.
Geometry-driven room layers struggle in open-plan spaces; KGSW seeds a watershed from
VLM-defined functions over wall-aware geometric cores.

\subsection{Small, fine-grained object retrieval and navigation}
\label{sec:rw-nav}
Map-based object-goal systems~\citep{Gu2024Conceptgraphs,Werby2024Hierarchical} and
FSR-VLN~\citep{Zhou2025Fsrvln} assume a mapped target or room cue. Zero-shot navigation
explores without a map~\citep{Zhou2023ESC,Yokoyama2024Vlfm,Yin2024Sgnav}, but is generally
coarse-grained. CARA accepts room-free fine-grained queries and escalates from map lookup
through 2D tags and VVC to active search, returning a coverage-conditioned $\varnothing$-decision.

\section{Method: \method{}}
\label{sec:method}

\subsection{Overview and notation}
\label{sec:overview}

\method{} builds an object-level open-vocabulary map from a posed RGB-D stream that
grounds two downstream capabilities. At frame $t$ the front-end (\S\ref{sec:e2g})
yields detections $\mathcal{D}^t=\{\detn_i\}$, each with a depth-back-projected 3D box
$\bbi_i$, a point set $P_i$, a normalized CLIP feature
$\cfeat_i\in\mathbb{R}^{d}$~\citep{Radford2021Learning}, and
an open-vocabulary label $\lbl_i$. An object node $\obj_j\in\mathcal{O}$ accumulates
points, a 3D box $\bbo_j$, a fused CLIP feature $\ofeat_j$, and a label history $C_j$
(the multiset of merged detection class ids); its majority label
$\lblstar_j=\operatorname{mode}(C_j)$ is both the object's final class and the
reference label LAMP (\S\ref{sec:ipa}) uses to test whether a detection matches it. Per
frame, IPA (\S\ref{sec:ipa}) associates each detection with an existing instance or
spawns a new one; the resulting high-fidelity map feeds KGSW room segmentation
(\S\ref{sec:kgsw}) and CARA retrieval (\S\ref{sec:cara}).

\subsection{E2G: Open-Vocabulary Perception with Selective VLM Invocation}
\label{sec:e2g}
\method{} performs open-vocabulary perception with an \emph{Enumerate-to-Ground} (E2G) front-end and FAV-Gate. For each RGB-D frame, E2G obtains visible object types from a VLM, FAV-Gate filters them into a stable, frame-specific vocabulary, and the grounder, SAM2, and CLIP produce the labeled detections, masks, and features used to build the map.

The VLM supplies three capabilities unavailable to cheaper on-device models (CLIP, SAM2, and the grounder): \emph{open-world naming}, \emph{fine-grained vision--language judgment}, and \emph{commonsense reasoning}. It therefore enumerates object types in E2G, confirms candidates and scores region priors in CARA (\S\ref{sec:cara}), and groups objects by function in KGSW (\S\ref{sec:kgsw}). Because cloud-VLM calls are expensive, \method{} handles routine cases with geometry and CLIP and reserves VLM calls for fine-grained semantic decisions (CARA's early tiers issue \emph{zero} VLM calls).

\paragraph{E2G: enumerate-to-ground perception.}
For each frame, E2G answers \emph{what} and \emph{where}: a VLM (Kimi-k2.6~\citep{KimiTeam2025KimiK2}) enumerates the visible object types $T^t$; LocateAnything~\citep{Wang2026LocateAnything} boxes each queried word, SAM2~\citep{Ravi2024SAM2} refines masks, and a CLIP encoder gives each detection a feature $\cfeat_i$ as a 2D prior---the one Tier~2 of CARA later uses (\S\ref{sec:cara}).

\paragraph{FAV-Gate (frame-adaptive vocabulary).}
$T^t$ is open but noisy---sporadic single-frame hallucinations fed to the grounder
cause false detections. FAV-Gate admits only words recurring in $\ge n_{\min}$ frames
(minus a tiny blacklist $\mathcal{B}$, e.g.\ text/logo/pipe) into a stable set
$V_{\mathrm{glob}}$, and queries each frame with their intersection:
\begin{equation}
\begin{aligned}
  V^t &= T^t \cap V_{\mathrm{glob}}, \\
  V_{\mathrm{glob}} &= \{\, w : \operatorname{freq}(w)\ge n_{\min}\,\}\setminus \mathcal{B},
\end{aligned}
\label{eq:fav}
\end{equation}
where $\operatorname{freq}(w)$ is the number of frames whose VLM tags contain word $w$.
The frequency floor removes enumeration noise and the intersection keeps each query
small and in-view, giving an \emph{open yet non-divergent} vocabulary.

\subsection{Instance-Preserving Object Update}
\label{sec:ipa}
Each detection is first associated with an instance and then used to update its
semantic feature.

\begin{figure}[t]
\centering
\makebox[\linewidth][c]{%
  \makebox[0.41\linewidth][c]{\textbf{(a) ConceptGraphs}}%
  \makebox[0.41\linewidth][c]{\textbf{(b) OVIP-SG (ours)}}}\par
\includegraphics[width=0.82\linewidth]{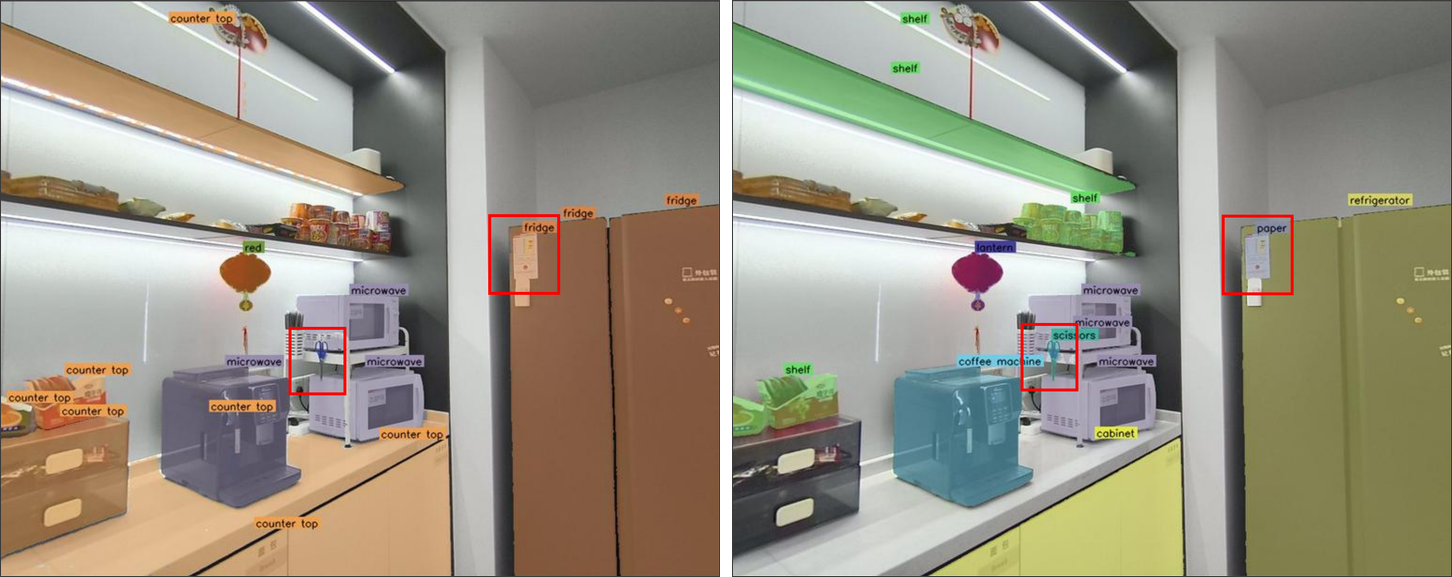}
\caption{Instance-preserving association. Containment association absorbs
fine-grained detections into enclosing objects, whereas IPSM+LAMP keeps attached
objects independent (red boxes).}
\label{fig:ipsm}
\end{figure}

\paragraph{Association: IPA (IPSM + LAMP).}
ConceptGraphs' containment score can be one when a small detection lies inside a
large object~\citep{Gu2024Conceptgraphs}. IPSM replaces it with symmetric 3D IoU, so
for $\bbi_i\subseteq\bbo_j$ the spatial score is
$\Vol(\bbi_i)/\Vol(\bbo_j)$ rather than one. LAMP additionally penalizes a mismatch
between detection label $\lbl_i$ and the object's majority label
$\lblstar_j=\operatorname{mode}(C_j)$:
\begin{equation}
\begin{aligned}
\aggm_{ij}={}&(1+\alpha)\frac{\Vol(\bbi_i\cap\bbo_j)}{\Vol(\bbi_i\cup\bbo_j)}
 +(1-\alpha)\cos(\cfeat_i,\ofeat_j)\\
&-w_L\Ind[\lbl_i\neq\operatorname{mode}(C_j)].
\end{aligned}
\label{eq:ipa}
\end{equation}
We assign $\detn_i$ to $j^\star=\arg\max_j\aggm_{ij}$ when
$\aggm_{ij^\star}>\tau$; otherwise, it starts a new object. Thus IPSM+LAMP protects
small attached instances and preserves the label purity needed for CARA's Tier~1
name lookup (\S\ref{sec:cara}). Fig.~\ref{fig:ipsm} contrasts containment, which
absorbs fine-grained detections into enclosing furniture, with IPSM+LAMP, which retains
independent instances; $w_L=0$ is the label-agnostic ablation.

\paragraph{View-area feature update.}
Equal-count fusion can overemphasize frequent small or peripheral crops. For a
detection $k$ with box area $a_k$ and center $\mathbf p_k$, we use the view-area weight
\begin{equation}
\begin{aligned}
w_k&=\frac{a_k}{HW}\exp\!\left[
-\frac{2\|\mathbf p_k-\mathbf p_0\|_2^2}{\sigma_v^2(W^2+H^2)}
\right],\\
\mathbf p_0&=(W/2,H/2),\qquad \sigma_v=0.6,
\end{aligned}
\label{eq:view-area-weight}
\end{equation}
This favors large, centered observations and discounts truncated or off-axis
crops~\citep{Jatavallabhula2023Conceptfusion}.
For observations assigned to $\obj_j$, online accumulators update as
\begin{equation}
\mathbf S_j\gets\mathbf S_j+w_k\cfeat_k,\quad
\Omega_j\gets\Omega_j+w_k,\quad
\ofeat_j=\frac{\mathbf S_j}{\|\mathbf S_j\|_2}.
\label{eq:awf}
\end{equation}
This is the normalized view-area--weighted mean computed without retaining historical
features; it updates only the associated object's semantic feature, not its geometry
or instance identity, and requires no additional perception model.

\subsection{KGSW: knowledge-guided semantic room segmentation}
\label{sec:kgsw}

KGSW converts the object map into functional floor regions. Geometry constrains each region to free space; semantics is used only to divide a geometrically connected open area when its objects imply different functions. We project free space and object centers to a bird's-eye-view (BEV) grid, with free-space mask $F$. For a free grid location $\mathbf x\in F$, the distance field $D(\mathbf x)=\min_{\mathbf y\notin F}\|\mathbf x-\mathbf y\|_2$ is the Euclidean clearance to the closest non-free grid location $\mathbf y$. Algorithm~\ref{alg:kgsw} uses this field to construct geometry-respecting cores and to expand the final semantic markers.

The VLM assigns each object $i$ a functional type $t_i$ from its semantic label and BEV center $\mathbf p_i$. This uses a soft commonsense relation between object labels and room functions, analogous to the soft commonsense constraints used by ESC for object navigation~\citep{Zhou2023ESC}. For a core $c$, let $\mathcal O_c=\{i:\mathbf p_i\in c\}$ be its object indices and $\mathcal T_c=\{t_i:i\in\mathcal O_c\}$ their functional types. An empty core with area at least $A_{\min}$ is retained as \texttt{unknown}; a core with one functional type remains one group. Only mixed cores are subdivided: objects are first separated by functional type and then by BEV proximity, so spatially disconnected areas with the same function do not share a marker. For an index set $I$, let $\mathsf{DB}_{\epsilon}(I)$ denote the DBSCAN partition of $\{\mathbf p_i:i\in I\}$, with neighborhood radius $\epsilon$~\citep{Ester1996DBSCAN}. KGSW therefore forms seed groups by
\begin{equation}
\Pi_c=
\begin{cases}
\{\mathcal O_c\}, & |\mathcal T_c|=1,\\
\displaystyle\biguplus_{t\in\mathcal T_c}
\mathsf{DB}_{\epsilon}\!\left(\{i\in\mathcal O_c:t_i=t\}\right),
& |\mathcal T_c|>1,
\end{cases}
\label{eq:kgsw-fusion}
\end{equation}
Each group supplies a marker disk of radius $r$ and its functional type to the final watershed. Thus, geometry determines coverage and boundaries, while VLM semantics resolves only the ambiguous mixed-region case; Algorithm~\ref{alg:kgsw} specifies the complete construction.

\begin{algorithm}[tb]
\caption{Knowledge-Guided Semantic Watershed (KGSW)}
\label{alg:kgsw}
\textbf{Input}: point cloud $P$; objects
$\mathcal O=\{(\lbl_i,\mathbf p_i)\}$; $\epsilon$, marker radius $r$, and minimum
core area $A_{\min}$\\
\textbf{Output}: room partition $R$ and floor--room graph $G$
\begin{algorithmic}[1]
\STATE $F\leftarrow\textsc{FreeSpace}(P)$;
$D(\mathbf x)\leftarrow\min_{\mathbf y\notin F}\|\mathbf x-\mathbf y\|_2$
\STATE $C_0\leftarrow\textsc{ConnectedComponents}
(F\setminus\textsc{Dilate}(\neg F))$
\STATE Remove components of $C_0$ below the core threshold
\STATE $C\leftarrow\textsc{Watershed}(-D,C_0;F)$
\STATE $\{t_i\}\leftarrow\textsc{VLMFunctionalGrouping}
(\{(\lbl_i,\mathbf p_i)\})$
\STATE Snap each $\mathbf p_i$ to free space; $M\leftarrow\varnothing$
\FOR{each geometric core $c\in C$}
  \STATE $\mathcal O_c\leftarrow\{i:\mathbf p_i\in c\}$
  \IF{$\mathcal O_c=\varnothing$}
    \IF{$\operatorname{Area}(c)\ge A_{\min}$}
      \STATE Add $c$ to $M$ with type \texttt{unknown}
    \ENDIF
  \ELSE
    \STATE $\mathcal T_c\leftarrow\{t_i:i\in\mathcal O_c\}$
    \IF{$|\mathcal T_c|=1$}
      \STATE $\Pi_c\leftarrow\{\mathcal O_c\}$
    \ELSE
      \STATE $\Pi_c\leftarrow\biguplus_{t\in\mathcal T_c}
      \mathsf{DB}_{\epsilon}(\{i\in\mathcal O_c:t_i=t\})$
    \ENDIF
    \FOR{each group $g\in\Pi_c$}
      \STATE Add radius-$r$ disks at $\{\mathbf p_i:i\in g\}$ to $M$
      \STATE Attach functional type $t(g)$ to the group
    \ENDFOR
  \ENDIF
\ENDFOR
\STATE $R\leftarrow\textsc{Watershed}(-D,M;F)$
\STATE $G\leftarrow\textsc{BackProjectFloorRoomGraph}(R,P)$
\STATE \textbf{return} $R,G$
\end{algorithmic}
\end{algorithm}

\subsection{CARA: cascaded active retrieval with absence assessment}
\label{sec:cara}

Given a free-form language query, CARA grounds it to a 3D goal and a furniture-aware
path through four tiers, from cheap/known to expensive/active
(Fig.~\ref{fig:overview}); each tier is invoked \emph{only} when cheaper evidence is
unavailable or rejected:
\begin{itemize}
\item \textbf{Tier~1 (map lookup):} the target matches a map object by class name or by
  the CLIP similarity of its class name; return that object's 3D centroid.
\item \textbf{Tier~2 (2D-tag retrieval):} the target is absent from the map but carries
  a 2D prior---it was \emph{named} by the per-frame VLM tags; CLIP scores the query
  over each frame's tag features, hot frames are bridged into segments, and a VLM
  disambiguates attributes on the best frame (e.g., which towel is floral).
\item \textbf{Tier~3 (history-frame VVC):} the target was never tagged; a VLM picks a
  likely region via the ESC commonsense prior
  $S(\text{goal}\mid r)$~\citep{Zhou2023ESC}, and a strong grounder re-run on that
  region's history frames is localized by multi-view voting (VVC).
\item \textbf{Tier~4 (active search + $\varnothing$-decision):} history frames also fail;
  the robot searches waypoints---sampled with clearance no smaller than the robot
  radius and excluding object footprints---rotating through six $60^\circ$ headings and
  moving along shortest paths on the navigation graph, with region order set by the
  schedule below; exhausting all regions yields a coverage-conditioned negative
  decision.
\end{itemize}

\paragraph{Region scheduling (semantics + distance).}
Within a region, CARA visits the nearest waypoint first; it selects the next region by
trading the VLM prior against normalized geodesic cost:
\begin{equation}
  r^{\star} = \operatorname*{arg\,max}_{r\in\mathcal{R}}\;
  \lambda\, S(\text{goal}\mid r)\;-\;(1-\lambda)\,\frac{g(p_{\text{cur}}, c_r)}{d_{\max}},
  \label{eq:region}
\end{equation}
Here $S(\text{goal}\mid r)$ is the VLM prior, $g$ is the distance to region centroid
$c_r$, and $\lambda$ balances semantic priority and proximity.

\paragraph{Trigger legitimacy.}
Each tier runs only after cheaper evidence is rejected. Tier~4 requires no accepted map
object, VLM tag, or history candidate; tagged targets are handled by Tier~2 rather than
active search.

\paragraph{Voxel-Vote Confirmation.}
Tiers~3--4 back-project detections into $0.25$\,m voxels, retain $\ge k$-frame
same-voxel consensus, and verify each survivor with a VLM crop. Voting preserves
fine-object recall; verification rejects consistent semantic false positives.

\paragraph{$\varnothing$-decision.}
After all coverage viewpoints are searched, if no sufficiently voted voxel is VLM
verified, CARA returns a coverage-conditioned negative decision rather than assuming
the target exists.

\paragraph{Cross-scene generality.}
The same cascade runs on ReplicaCAD and HM3D in Habitat and is deployed online and in
closed loop in office and hotel environments (\S\ref{sec:main-exp}).

\section{Experiments}
\label{sec:main-exp}

\paragraph{Protocol.}
We evaluate Task~A mapping on six Replica scenes. Replica is a widely used simulated
indoor benchmark for semantic mapping and provides dense semantic and instance-level
annotations. We use one shared ConceptGraphs evaluator: every method is reclassified
over the same 52 classes with the same CLIP template, matched to GT by $k{=}1$ nearest
neighbor, and averaged per scene. For instance evaluation, we use native Replica mesh
instances derived from official face-level object IDs and report class-agnostic
PQ/SQ/RQ at IoU$>0.5$. KGSW is evaluated on three HM3D scenes with the HOV-SG room
evaluator. HM3D comprises scans captured from real indoor environments and therefore
complements Replica with realistic geometry and layouts. Task~B uses ReplicaCAD and
two HM3D scenes; independent semantic-mesh GT is used only for scoring. Full per-scene
results, sensitivity analyses, query definitions, and hyperparameters are in
the supplementary material.

\subsection{Task A: Mapping Fidelity}

\begin{figure*}[t]
\centering
\includegraphics[width=\textwidth]{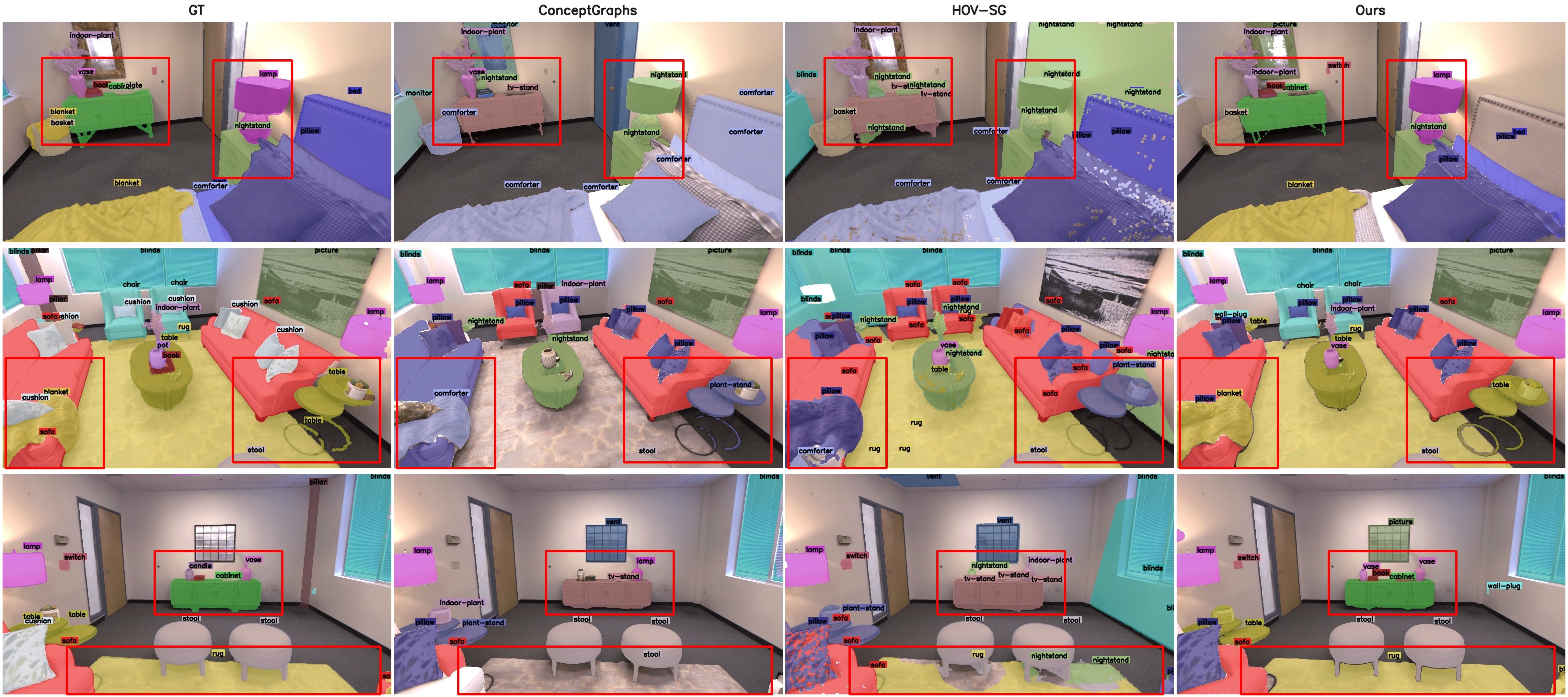}
\caption{Replica qualitative comparison under the same projected-label protocol.
ConceptGraphs misses long-tail objects, while HOV-SG fragments open surfaces.
OVIP-SG preserves small instances (e.g., switch, wall-plug, picture, vase) and
keeps large furniture intact; red boxes mark the differences.}
\label{fig:qualcompare-main}
\end{figure*}

\paragraph{Accuracy and attribution.}
Under the shared protocol, OVIP-SG improves ConceptGraphs from 45.00 to
51.31 mAcc and from 56.30 to 61.45 F-mIoU (Table~\ref{tab:main-main}). It leads
mAcc on five of six scenes; the room2 regression is reported in
the supplementary material. The cumulative ablation fixes one E2G detection
cache for all rows after the baseline (Table~\ref{tab:stack-main}). The
front-end swap contributes the largest mAcc gain ($+3.99$); IPSM, LAMP, and
AWF add $+2.32$ on identical detections. Class-agnostic native-instance PQ is 0.398,
versus 0.338 for ConceptGraphs and 0.163 for HOV-SG
(Table~\ref{tab:instance-main}); OVIP-SG also leads SQ/RQ (0.776/0.514) and has the
highest PQ on five of six scenes. The size-stratified results
(Table~\ref{tab:instance-size}) show gains for small and large instances. Targeted swallowed-class recall rises
from 15.8 to 43.2 with IPSM+LAMP. With association fixed, AWF raises scene-mean F-mIoU
$50.09\!\to\!61.45$ (mAcc: $50.35\!\to\!51.31$; see the supplementary material). The qualitative
comparison in Fig.~\ref{fig:qualcompare-main} further shows that OVIP-SG preserves
small instances while keeping large furniture intact.

\begin{table}[t]
\centering
\begin{tabular}{lrr}
\toprule
Method & mAcc$\uparrow$ & F-mIoU$\uparrow$ \\
\midrule
ConceptFusion & 35.63 & 47.95 \\
HOV-SG & 40.98 & 58.73 \\
ConceptGraphs & 45.00 & 56.30 \\
\textbf{OVIP-SG (Ours)} & \textbf{51.31} & \textbf{61.45} \\
\bottomrule
\end{tabular}
\caption{Replica mapping, scene mean under one evaluator.}
\label{tab:main-main}
{\small
\begin{tabular}{lcc}
\toprule
Configuration & mAcc & $\Delta$ \\
\midrule
ConceptGraphs (RAM) & 45.00 & --- \\
E2G + overlap & 48.99 & $+3.99$ \\
$\quad$+ IPSM & 49.42 & $+0.43$ \\
$\quad$+ LAMP & 50.35 & $+0.93$ \\
$\quad$+ AWF & \textbf{51.31} & $+0.96$ \\
\bottomrule
\end{tabular}
}
\caption{Cumulative mAcc ablation; rows after the baseline share detections.}
\label{tab:stack-main}
\end{table}

\paragraph{VLM-backbone robustness.}
We repeat the final mapping pipeline while changing only the stage-1 VLM used for
per-frame enumeration; the stable vocabulary, LocateAnything grounding, SAM2$+$CLIP
features, ConceptGraphs mapping backend, area-weighted fusion, and evaluator are
held fixed. For a strict four-backbone comparison, Table~\ref{tab:vlm_ablation}
uses the five scenes available for every backbone (room0/1/2 and office0/2).
The $1.71$-point mAcc and $2.39$-point F-mIoU ranges show that the result is robust
to this choice: GPT-5.5 and Sonnet-5 are modestly stronger, whereas the Kimi and
Qwen variants remain in the same narrow band. This is expected because the VLM
enumerates candidates, while the final semantic features come from area-weighted CLIP
features on SAM2 crops.

\begin{table}[t]
\centering
{\small
\setlength{\tabcolsep}{2mm}
\begin{tabular}{lcc}
\toprule
VLM backbone & mAcc$\uparrow$ & F-mIoU$\uparrow$ \\
\midrule
GPT-5.5            & \textbf{53.31} & \textbf{61.73} \\
Claude-Sonnet-5    & 52.81 & 60.97 \\
Kimi-K2.6 (base)   & 51.73 & 59.63 \\
Qwen3.7-Plus       & 51.60 & 59.34 \\
\bottomrule
\end{tabular}
}
\caption{VLM-backbone ablation for Task~A mapping on Replica. All backbones share
the identical pipeline; only the stage-1 per-frame VLM enumeration is swapped.
Scene mean over the five scenes shared by all backbones (room0/1/2, office0/2).
mAcc / F-mIoU (\%).}
\label{tab:vlm_ablation}
\end{table}

\begin{table}[t]
\centering
{\small
\begin{tabular}{lccc}
\toprule
Method & PQ$\uparrow$ & SQ$\uparrow$ & RQ$\uparrow$ \\
\midrule
HOV-SG                    & 0.163 & 0.721 & 0.227 \\
ConceptGraphs             & 0.338 & 0.740 & 0.461 \\
\textbf{OVIP-SG (Ours)}  & \textbf{0.398} & \textbf{0.776} & \textbf{0.514} \\
\bottomrule
\end{tabular}
}
\caption{Class-agnostic instance quality against native Replica mesh instances
(six-scene mean; official face-level object IDs propagated to vertices by
area-weighted incident-face voting, IoU$>0.5$). ConceptFusion is omitted because its
dense point field has no semantic instance units.}
\label{tab:instance-main}
{\small
\begin{tabular}{lccc}
\toprule
Method & Small$\uparrow$ & Medium$\uparrow$ & Large$\uparrow$ \\
\midrule
HOV-SG & 0.213 & 0.423 & 0.624 \\
ConceptGraphs & 0.164 & \textbf{0.587} & 0.619 \\
\textbf{OVIP-SG (Ours)} & \textbf{0.254} & 0.582 & \textbf{0.712} \\
\bottomrule
\end{tabular}
}
\caption{Native-instance recall at IoU$>0.5$ by per-scene GT mesh surface-area
tercile (small/medium/large), averaged over six Replica scenes.}
\label{tab:instance-size}
{\small
\setlength{\tabcolsep}{1mm}
\textit{Room-GT segmentation / oracle search unit}\par
\begin{tabular}{@{}lcccc@{}}
\toprule
Method & HydP$\uparrow$ & acc@0.5$\uparrow$ & AP$\uparrow$ & Area$\downarrow$ \\
\midrule
HOV-SG & \textbf{0.809} & 0.567 & 0.567 & 0.342 \\
SysNav & 0.529 & 0.351 & 0.378 & 0.713 \\
\textbf{KGSW (Ours)} & 0.791 & \textbf{0.593} & \textbf{0.698} & \textbf{0.218} \\
\bottomrule
\end{tabular}
}
\caption{KGSW functional-region segmentation and oracle search evidence on three
HM3D scenes (217 queries). Area is the oracle target-containing region's floor-area
fraction.}
\label{tab:kgsw-main}
\end{table}

\paragraph{Functional regions.}
Geometry-only room partitions collapse in elongated or half-open layouts:
SysNav falls to 0.351 mean acc@0.5 (Table~\ref{tab:kgsw-main}).
KGSW remains stable, attaining the
best mean acc@0.5 (0.593), AP (0.698), and lowest scene variance. Its purpose is
functional search partitioning rather than strict room-GT recovery: over 217
GT-object queries, the oracle target-containing region occupies 0.218 of the floor,
$1.6\times$ smaller than HOV-SG and $3.3\times$ smaller than SysNav, at similar
containment (Table~\ref{tab:kgsw-main}). The corresponding open-area subdivision is shown in
Fig.~\ref{fig:roomseg}; the single-scene breakdown is provided in
the supplementary material.

\begin{figure}[t]
\centering
\includegraphics[width=0.98\columnwidth]{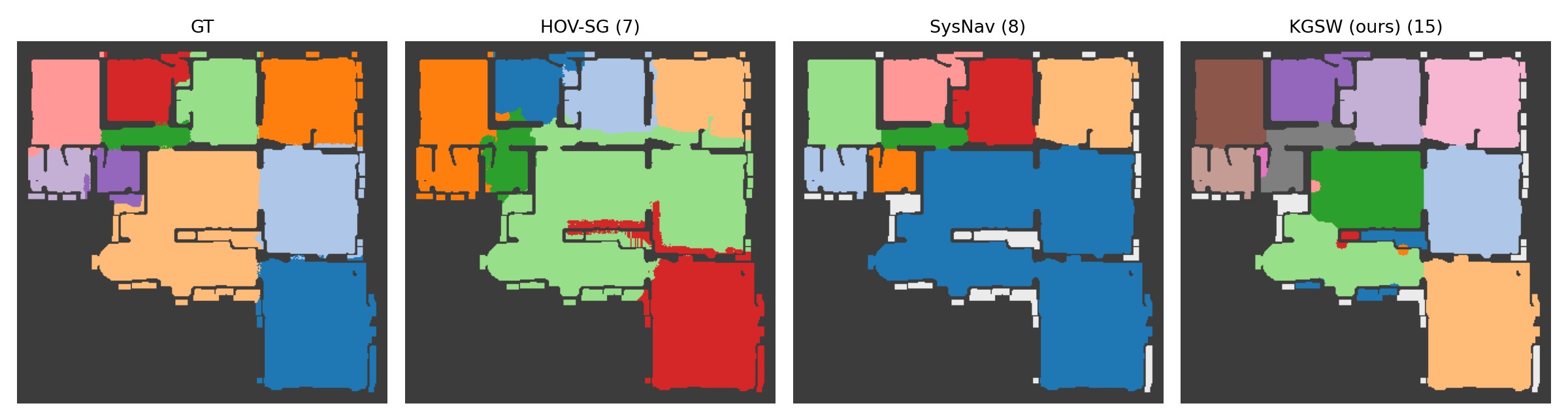}
\caption{Top-down room segmentation on an HM3D scene. HOV-SG and the SysNav-style
geometric watershed preserve walled rooms but merge the open area; KGSW divides it
into functional regions.}
\label{fig:roomseg}
\end{figure}

\subsection{Task B: Small, Fine-Grained Object Retrieval and Navigation}

\paragraph{Cascade coverage.}
CARA escalates only when cheaper evidence fails. Across ReplicaCAD, HM3D, and
real-robot office and hotel deployments, its four tiers cover mapped,
tagged-but-unmapped, history-only, and active-search/negative cases.
Table~\ref{tab:cara-main} lists one representative case per tier---a tagged floral
towel, a history-only coffee machine and robot-recorded noodles, an actively found
knife holder, and an absent piano; the full inventory is in
the supplementary material.

\begin{table}[t]
\centering
\setlength{\tabcolsep}{1mm}
\begin{tabular}{@{}l >{\raggedright\arraybackslash}p{0.24\linewidth} >{\raggedright\arraybackslash}p{0.50\linewidth}@{}}
\toprule
Tier & Capability & Representative evidence \\
\midrule
T1 & mapped-object lookup     & 6 targets; apt\_2 goals reached within $0.25$--$0.45$\,m \\
T2 & tagged off-map retrieval & floral-towel disambiguation; 3 apt\_2 targets \\
T3 & history-only recovery    & coffee machine: 8 votes, $0.06$\,m; real-robot noodles \\
T4 & active search / absence  & knife holder at $8/52$; piano absent after $52/52$ views \\
\bottomrule
\end{tabular}
\caption{Representative CARA capability cases, one per tier. These are
\emph{representative capability cases, not an aggregate benchmark}; the full
inventory is in the supplementary material.}
\label{tab:cara-main}
\centering
\setlength{\tabcolsep}{0.35mm}
\begin{tabular}{@{}lcccccc@{}}
\toprule
 & \multicolumn{3}{c}{Top-1 RSR$@d$} & \multicolumn{3}{c}{Top-5 RSR$@d$} \\
\cmidrule(lr){2-4}\cmidrule(lr){5-7}
Method [source] & 1m & 2m & 3m & 1m & 2m & 3m \\
\midrule
HOV-SG (paper) & 0.52 & 0.64 & \textbf{0.70} & \textbf{0.76} & 0.82 & 0.88 \\
HOV-SG (reprod.) & 0.51 & \textbf{0.66} & \textbf{0.70} & 0.70 & 0.82 & 0.85 \\
FSR-VLN (open src.) & 0.49 & 0.57 & 0.67 & 0.69 & 0.79 & 0.82 \\
\textbf{CARA (Ours)} & \textbf{0.57} & 0.61 & 0.63 & 0.73 & \textbf{0.88} & \textbf{0.93} \\
\bottomrule
\end{tabular}
\caption{HM3D object-retrieval success rate (RSR): paper and reproduced results;
paper sources are \citet{Werby2024Hierarchical} and \citet{Zhou2025Fsrvln}.}
\label{tab:rsr}
\centering
\setlength{\tabcolsep}{0.5mm}
\begin{tabular}{@{}lrrr@{}}
\toprule
Goal policy & SR$\uparrow$ & SPL$\uparrow$ & Path (m) \\
\midrule
B-random  & $.136{\pm}.000$ & $.087{\pm}.000$ & 13.32 \\
B-nearest & $.091{\pm}.000$ & $.091{\pm}.000$ & 0.55  \\
B-lexical & $.485{\pm}.021$ & $.248{\pm}.002$ & 14.00 \\
B-CLIP    & $.576{\pm}.021$ & $\mathbf{.384{\pm}.020}$ & 11.29 \\
\textbf{CARA (Ours)} & $\mathbf{.605{\pm}.011}$ & $.339{\pm}.002$ & 14.02 \\
\bottomrule
\end{tabular}
\caption{Closed-loop HM3D simulation (22 present queries, mean$\pm$std over three
starts; shared map, starts, follower, and evaluator). Only CARA supports a searched
$\varnothing$-decision.}
\label{tab:e9-main}
\end{table}

\begin{table}[!t]
\centering
\setlength{\tabcolsep}{0.5mm}
\begin{tabular}{@{}lrrrrrrrr@{}}
\toprule
 & \multicolumn{4}{c}{Strict counts} & \multicolumn{4}{c}{Rates} \\
\cmidrule(lr){2-5}\cmidrule(lr){6-9}
Budget & TP & FN & TN & FP & Sens.$\uparrow$ & Abs.$\uparrow$ & Bal.$\uparrow$ & Loc.$\uparrow$ \\
\midrule
Full & 20 & 2 & 14 & 8 & .909 & .636 & .773 & .455 \\
Half & 19 & 3 & 14 & 8 & .864 & .636 & .750 & .409 \\
\bottomrule
\end{tabular}
\caption{Tier-4 stress test on two HM3D scenes. Half is the exact 50\% trace prefix;
strict counts retain eight visually disputed FPs.}
\label{tab:absence-main}
\end{table}

\begin{figure}[t]
\centering
\includegraphics[width=0.95\linewidth]{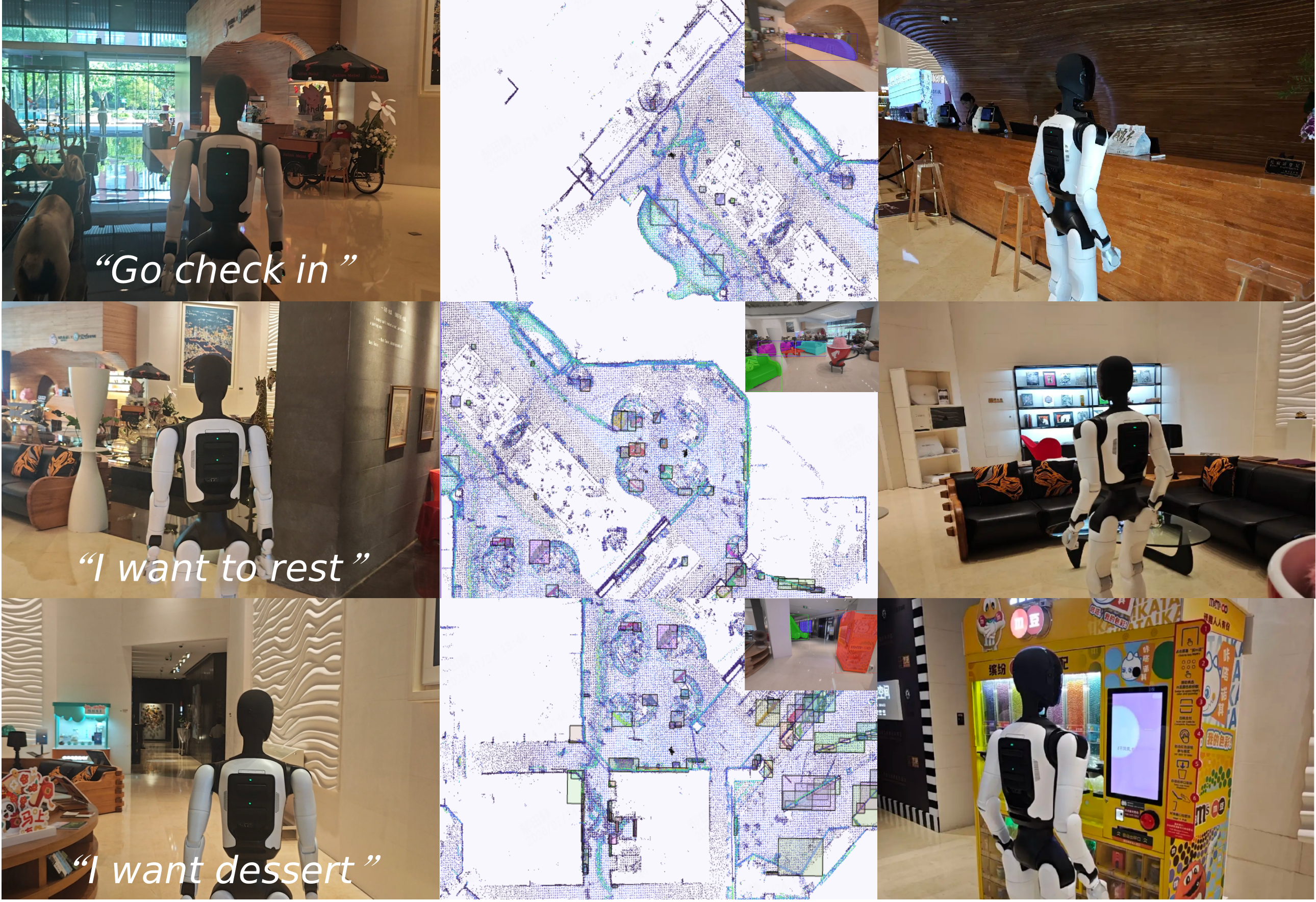}
\caption{Qualitative real-world executions of language-guided service
requests with the full map+CARA stack. The three rows show check-in, rest, and dessert
requests. Left and right show the initial and
destination views; the middle shows the global LiDAR map overlaid with RGB semantic
point cloud and 3D instance bounding boxes.}
\label{fig:realrobot}
\end{figure}

Under the shared open-source HM3D retrieval scorer, HOV-SG is reproduced at 0.51
(paper: 0.52), FSR-VLN at 0.49, and CARA at 0.57 (Top-5@3\,m: 0.93). Its added
contribution is the Tiers~2--4 capability surface (Table~\ref{tab:rsr}).

\paragraph{Closed-loop HM3D navigation.}
In Table~\ref{tab:e9-main}, each policy predicts its goal and GT only scores it. CARA has higher mean
SR than open-vocabulary CLIP ($0.605\pm0.011$ vs. $0.576\pm0.021$).
Its added capability is a searched
negative decision, unavailable to the map-only baselines. Audits verify identical
starts, non-degenerate SPL, and non-GT CARA goals; failures mainly reflect map-goal
error rather than solved low-level navigation.

\paragraph{Balanced Tier-4 negative-decision test.}
We force all 44 semantic-mesh-balanced queries (22 present/22 absent; two scenes)
through Tier~4, so this is a controlled module test, not the cascade's natural
trigger rate. Full coverage yields TP/FN/TN/FP $=20/2/14/8$: sensitivity is
$0.909$ (Wilson 95\% CI $[0.722,0.975]$), strict annotation-referenced absent
accuracy is $0.636$ ($[0.430,0.803]$), and balanced accuracy is $0.773$
(Table~\ref{tab:absence-main}); Loc@1m is separately $10/22$. We retain all eight
strict FPs---each crop contains target-like content---and apply no post-hoc
correction. Thus $\varnothing$ remains conditioned on both searched coverage and
annotation completeness.

\subsection{Real-Robot Deployment}

We deploy the complete map+CARA stack online and in closed loop on an AgiBot
Expedition A3 humanoid robot in office and hotel environments with distinct layouts and
service-relevant objects. Using the synchronized sensor streams, the robot incrementally
builds its semantic map, grounds each free-form service request, and navigates to the
resulting 3D goal during live execution; the 3D goal is executed by the A3's native
navigation and motion-control stack.

The task set spans large-object retrieval, small-object retrieval, and goal-navigation
at increasing difficulty across both environments. We report Success Rate (SR), the fraction of episodes completed successfully. On
small-object retrieval tasks, OVIP-SG improves SR over FSR-VLN by 0.08 on average.
Fig.~\ref{fig:realrobot} shows representative successful executions.

\section{Conclusion}
\label{sec:conclusion}
We presented \method{}, an instance-preserving open-vocabulary object map for
functional segmentation, retrieval, navigation, and coverage-conditioned negative
decisions. At mapping time, symmetric association and view-area fusion preserve small
independent instances; at query time, KGSW and CARA turn the map into functional priors
and a coverage-aware search policy. Under one evaluator it improves mapping accuracy and
instance quality over ConceptGraphs; KGSW yields stable functional regions; and CARA attains
higher retrieval SR than the strongest baseline with an explicit, coverage-conditioned
negative decision. The same representation supports free-form service requests on the real
robot, while the Tier-4 test makes searched absence an explicit outcome. Limitations
include real-world scale and annotation fidelity.

\clearpage
\makeatletter
\@fptop=0pt plus 1fil
\makeatother

\clearpage
\appendix
\section{Full Experimental Details}
\label{sec:exp}

\begin{table}[b]
\centering
{\small
\setlength{\tabcolsep}{1mm}
\begin{tabular}{@{}l >{\raggedright\arraybackslash}p{0.72\columnwidth}@{}}
\toprule
Acronym & Meaning \\
\midrule
E2G & Enumerate-to-Ground \\
FAV-Gate & Frame-Adaptive Vocabulary Gate \\
IPA & Instance-Preserving Association \\
IPSM & Instance-Preserving Symmetric-IoU Merge \\
LAMP & Label-Aware Merge Penalty \\
AWF & Area-Weighted Fusion \\
KGSW & Knowledge-Guided Semantic Watershed \\
CARA & Cascaded Active Retrieval with Absence Assessment \\
VVC & Voxel-Vote Confirmation \\
$\varnothing$-decision & Tier-4 coverage-conditioned negative decision \\
\bottomrule
\end{tabular}
}
\caption{Component acronym reference.}
\label{tab:acronyms}
\end{table}

\subsection{Setup and evaluation protocol}
\label{sec:setup}
\paragraph{Datasets and metrics.}
We evaluate semantic mapping on Replica, a widely used simulated indoor benchmark for
semantic mapping, on the six scenes \{room0, room1, room2, office0, office2, office3\},
reporting class-mean accuracy (mAcc) and frequency-weighted mIoU (F-mIoU). We use HM3D,
which consists of scans captured from real indoor environments, to complement Replica
with realistic geometry and layouts for functional-region and retrieval evaluation.

\paragraph{Unified evaluation protocol.}
For a fair comparison, every method is scored by the same ConceptGraphs evaluation
routine: each object's fused CLIP feature is reclassified over the 52 Replica classes
with a shared text template (\texttt{"A photo of a \{label\}."}), matched to the GT
point cloud by nearest neighbor ($k{=}1$), excluding six structural/unlabeled classes.
We re-run ConceptFusion and, where feasible, HOV-SG through this same routine rather
than quoting their papers' numbers, so all systems share one classifier, one template,
and one GT alignment---replacing ConceptGraphs' weak default prompt
(\texttt{"an image of \{c\}"}) with CLIP's recommended
\texttt{"A photo of a \{label\}."}~\citep{Radford2021Learning} for all methods.

\paragraph{Aggregation protocol.}
We aggregate per scene and average over scenes (\emph{scene-mean}), the protocol used
by ConceptGraphs, ConceptFusion, and HOV-SG.

\paragraph{Native instance protocol.}
Replica's official semantic mesh stores an \texttt{object\_id} on each face. We
propagate face IDs to incident mesh vertices by area-weighted voting; if a retained
face ID receives no winning vertex, its incident vertices are assigned to it so no
official instance is discarded. We retain each non-void object ID as one GT instance,
and exclude floor, wall, ceiling, door, and window instances. The native mesh vertices
are the common support: each vertex is
assigned the object ID of its nearest predicted map point. We then compute
class-agnostic panoptic quality (PQ), segmentation quality (SQ), and recognition
quality (RQ) using greedy one-to-one matches at IoU$>0.5$. All prediction maps are
already in the mesh coordinate frame: across the 18 method--scene pairs, median
GT-vertex-to-prediction nearest-neighbor distances are 0.8--11.6\,cm, so no pose
transform is applied. ConceptFusion is excluded from this instance evaluation because
its dense point field has no semantic instance units.

\subsection{Task A: High-Fidelity Semantic Mapping}
\label{sec:exp-taskA}

\subsubsection{Mapping accuracy}
\label{sec:exp-mapping}
Table~\ref{tab:main} reports scene-mean accuracy over the six scenes, every
method---ours, ConceptGraphs, HOV-SG, and ConceptFusion---scored through the identical
pipeline. Our final model leads on \emph{both} metrics over the ConceptGraphs baseline:
mAcc $45.00\!\to\!51.31$ ($+6.31$) and F-mIoU $56.30\!\to\!61.45$ ($+5.15$). Per scene
(Table~\ref{tab:perscene}), our final model has the highest mAcc on five of the six
scenes; on room2 it dips below the baseline ($47.25$ vs.\ $51.14$), a documented
side-effect of area-weighted fusion (\S\ref{sec:exp-awf}, Table~\ref{tab:awf}) that
trades a little equally-weighted recall for large-object purity. HOV-SG,
state-of-the-art under its own looser protocol
($101$ classes, $k{=}5$ NN vote), falls to $40.98$ mAcc once re-scored through the
shared $52$-class $k{=}1$ protocol---below the baseline---while its frequency-weighted
F-mIoU ($58.73$) surpasses the baseline yet still trails ours: its DBSCAN
dominant-cluster feature is strong on large surfaces but its class-mean recall on the
long tail is weak. ConceptFusion, a per-point (non-instance) feature field, trails all
object-level methods. The AWF component driving the F-mIoU gain is isolated in
\S\ref{sec:exp-awf}.

\begin{table}[t]
\centering
\begin{tabular}{lrr}
\toprule
Method & mAcc & F-mIoU \\
\midrule
ConceptFusion & 35.63 & 47.95 \\
HOV-SG & 40.98 & 58.73 \\
ConceptGraphs & 45.00 & 56.30 \\
\midrule
\textbf{OVIP-SG (Ours)} & \textbf{51.31} & \textbf{61.45} \\
\bottomrule
\end{tabular}
\caption{Replica semantic mapping accuracy (scene-mean over 6 scenes; identical
\texttt{eval\_replica} pipeline and text template for all methods).}
\label{tab:main}
\end{table}

\begin{table}[t]
\centering
{\small
\setlength{\tabcolsep}{1mm}
\begin{tabular}{@{}lcccccc@{}}
\toprule
Method & r0 & r1 & r2 & o0 & o2 & o3 \\
\midrule
ConceptFusion & 41.24 & 50.62 & 38.29 & 32.17 & 28.39 & 23.05 \\
HOV-SG        & 49.14 & 49.11 & 36.19 & 37.93 & 41.19 & 32.30 \\
ConceptGraphs & 51.61 & 50.90 & \textbf{51.14} & 37.63 & 41.57 & 37.17 \\
\textbf{OVIP-SG (Ours)} & \textbf{58.22} & \textbf{63.25} & 47.25 & \textbf{42.62} & \textbf{47.30} & \textbf{49.22} \\
\bottomrule
\end{tabular}
}
\caption{Per-scene mAcc (same protocol as Table~\ref{tab:main}); r/o abbreviate
room/office. Our model is highest on five of six scenes; room2 is the AWF side-effect
discussed in \S\ref{sec:exp-awf}.}
\label{tab:perscene}
\end{table}

\subsubsection{Ablation: area-weighted fusion}
\label{sec:exp-awf}
Table~\ref{tab:awf} isolates AWF by re-fusing the same map with area weighting
(association unchanged). F-mIoU improves on five of six scenes and is flat on the
sixth, and mAcc rises slightly. The largest gains are on scenes where a large surface
is co-detected with many small parts (office2, room0). The single regression is room2
mAcc ($-4.68$): area weighting lets a few small classes yield to the dominant body,
lowering the equally-weighted mAcc but barely moving the frequency-weighted F-mIoU
($+0.24$)---the expected trade-off of favoring the visual subject.

\begin{table}[t]
\centering
\begin{tabular}{lrrrr}
\toprule
 & \multicolumn{2}{c}{mAcc} & \multicolumn{2}{c}{F-mIoU} \\
\cmidrule(lr){2-3}\cmidrule(lr){4-5}
Scene & base & AWF & base & AWF \\
\midrule
room0   & 53.59 & 58.22 & 58.86 & 79.93 \\
room1   & 58.67 & 63.25 & 44.80 & 53.11 \\
room2   & 51.93 & 47.25 & 71.98 & 72.22 \\
office0 & 43.71 & 42.62 & 37.98 & 37.55 \\
office2 & 48.42 & 47.30 & 22.63 & 55.36 \\
office3 & 45.80 & 49.22 & 64.27 & 70.52 \\
\midrule
\textbf{scene-mean} & 50.35 & \textbf{51.31} & 50.09 & \textbf{61.45} \\
\bottomrule
\end{tabular}
\caption{Per-scene effect of AWF on our method (association fixed; ``base'' = the
conventional detection-count running-mean fusion, ``AWF'' = our area-weighted fusion).}
\label{tab:awf}
\end{table}

\subsubsection{Ablation: association and front-end}
\label{sec:exp-assoc}
Figure~3 in the main paper visualizes the failure mode that matters for IPA: small or
attached objects can be absorbed into a larger instance. Whole-scene F-mIoU is not the
right attribution metric here, as it is dominated by large furniture and penalizes the
intentional reduction in over-merging. We therefore evaluate association with a
targeted recovery diagnostic: for classes visibly or quantitatively swallowed by the
pre-IPA map, we measure their label-area recall before and after the label-aware
penalty, keeping the detector, masks, features, and IPSM spatial term fixed.

The diagnostic matches the qualitative observation: when a target class is fully
absorbed (pot in room0, vase in room1) IPA recovers it from zero recall; when the
pre-IPA map already recovers it (room2 vase) the gain is small. We thus use IPA for
instance fidelity and downstream retrieval, and treat global mAcc/F-mIoU elsewhere only
as scene-level segmentation metrics, not as evidence for small-object preservation.

\paragraph{Same-detection module attribution.}
To attribute the $+6.31$ mAcc gain over the ConceptGraphs baseline, we hold the
\emph{identical} E2G detection cache fixed and toggle only the association
(overlap$\to$IPSM), LAMP, and AWF, re-scoring each with the shared protocol
(Table~\ref{tab:stack}). Two facts stand out and we state them plainly. First, the
open-vocabulary front-end is the largest single contributor: swapping the RAM front-end
for E2G (same overlap association and count-mean fusion) already lifts mAcc
$45.00\to48.99$ ($+3.99$). Second, on \emph{identical} detections the association/fusion
modules add a further $+2.32$: IPSM $+0.43$, LAMP $+0.93$, AWF $+0.96$. IPSM's global-mAcc
effect is small and scene-dependent (e.g.\ office0 $+7.8$ but room0/room1 lower); its
intended value is instance preservation; the cross-method native-instance evidence is
reported in the main paper, and AWF's value is F-mIoU
(Table~\ref{tab:awf}), neither of which global mAcc captures. We therefore do not
attribute the headline gain to association/fusion alone---the front-end drives most of the
mAcc, while IPA and AWF contribute instance fidelity and frequency-weighted quality
respectively.

\begin{table}[t]
\centering
{\small
\setlength{\tabcolsep}{1mm}
\begin{tabular}{@{}>{\raggedright\arraybackslash}p{0.70\columnwidth}cc@{}}
\toprule
Configuration & mAcc & $\Delta$ \\
\midrule
ConceptGraphs baseline (RAM front-end) & 45.00 & --- \\
E2G front-end + overlap (no IPSM/LAMP/AWF) & 48.99 & $+3.99$ \\
\quad + IPSM (symmetric 3D-IoU) & 49.42 & $+0.43$ \\
\quad + LAMP ($w_L{=}0.5$) & 50.35 & $+0.93$ \\
\quad + AWF (final) & \textbf{51.31} & $+0.96$ \\
\bottomrule
\end{tabular}
}
\caption{Same-detection cumulative ablation (scene-mean mAcc, shared protocol). Rows
below the baseline share one E2G (KIMI+LA) detection cache. The RAM$\to$E2G front-end
swap is the largest contributor; IPSM/LAMP/AWF add $+2.32$ on top.}
\label{tab:stack}
\end{table}

\subsubsection{Analysis and findings}
\label{sec:exp-analysis}
\paragraph{Where the gains come from.}
The gain is driven by long-tail small-object classes the closed-set baseline misses
entirely---wall-plug ($21.0\!\to\!81.1$), tissue-paper ($25.5\!\to\!76.2$), vent
($33.5\!\to\!70.5$), bed ($74.0\!\to\!99.8$)---which open-vocabulary enumeration
recovers as independent instances rather than absorbing into adjacent furniture. Yet
the final model is not a long-tail specialist that sacrifices common classes: it ranks
first on \emph{both} mAcc (favoring small classes) and F-mIoU (dominated by large
surfaces), so recovering the tail costs no accuracy on point-dense furniture.

\paragraph{Trusting the detector's own labels.}
The shared protocol reclassifies each object's feature over the 52 fixed classes and
\emph{discards the open-vocabulary label the detector already assigned}: on office2 the
detector correctly names a \texttt{sofa}, yet the CLIP text classifier flips it to the
near-synonym \texttt{cushion} by a $0.005$ cosine margin and marks it wrong. As a
supplementary diagnostic we report a \emph{label-weighted} evaluation mixing the CLIP
score with an area-weighted prior from the object's own detection label,
$s=(1-\lambda)\,s^{\mathrm{clip}}+\lambda\,s^{\mathrm{label}}$
(Table~\ref{tab:lambda}): including detector labels raises mAcc $51.31\!\to\!64.57$ and
F-mIoU $61.45\!\to\!79.99$. This is a diagnostic, not our cross-method ranking---it uses
labels not every baseline emits in the same form, so the main table stays at
$\lambda{=}0$---but the size of the gap shows how much reliable semantics a
pure-feature closed-set protocol discards.

\begin{table}[t]
\centering
\begin{tabular}{lrrr}
\toprule
$\lambda$ & mAcc & F-mIoU & mIoU \\
\midrule
0.0 (CLIP-only, main table) & 51.31 & 61.45 & 37.46 \\
0.3 (balanced) & \textbf{64.57} & 78.64 & \textbf{52.41} \\
1.0 (label-only) & 62.98 & \textbf{79.99} & 49.18 \\
\bottomrule
\end{tabular}
\caption{Label-weighted evaluation on our final maps (scene-mean). $\lambda{=}0$ is the
main table's CLIP-only protocol; larger $\lambda$ admits the detector's own
open-vocabulary label. Supplementary diagnostic, not used for cross-method ranking.}
\label{tab:lambda}
\end{table}

\subsection{Functional-Region Segmentation with KGSW}
\label{sec:exp-kgsw}
KGSW converts the object map into a floor--room graph. Its target is not classical
room segmentation but \emph{functional-region} segmentation: it deliberately subdivides
a large open space into semantically coherent search units (cooking vs.\ dining vs.\
lounge corner), which is what the downstream retriever needs. We nonetheless report it
under the HOV-SG HM3D evaluator---which matches predicted regions to annotated-room GT
by IoU and reports Hydra precision (HydP), acc@IoU=0.5, and AP---as a sanity check that
KGSW's regions are clean, while noting that a room-GT metric penalizes intentional
over-segmentation by construction. \#R is the predicted room/region count.

As visualized in Figure~5 of the main paper, KGSW subdivides the open area by function and
attains the highest Hydra precision ($0.954$ vs.\ HOV-SG $0.841$, SysNav $0.911$):
every predicted region is clean, i.e.\ contained in a single GT room. Its lower
acc@$0.5$ ($0.667$ vs.\ SysNav $0.800$) is the expected cost of targeting functional
regions rather than annotated rooms: acc@0.5 rewards a strict one-to-one match to the
ten annotated rooms, so methods with region counts near ten are favored, whereas KGSW
deliberately subdivides a large open area into several functional regions ($15$
predicted), leaving extra unmatched regions under a room-GT metric even though each
region is clean (hence the top HydP). This is a design choice, quantified---not
explained away---by the search-utility results below (Table~\ref{tab:kgsw_search_utility}):
splitting an open space by function (cooking vs.\ dining vs.\ lounge corner) gives the
robot smaller,
semantically coherent search units, so a language query can be resolved within the most
likely functional region---directly benefiting CARA's region-scoped retrieval below.

\begin{table}[t]
\centering
\begin{tabular}{@{}lrrr@{}}
\toprule
Method & \#R & HydP & acc@0.5 / AP \\
\midrule
HOV-SG & 7 & 0.841 & 0.700 \\
SysNav & 8 & 0.911 & \textbf{0.800} \\
\textbf{KGSW (Ours)} & \textbf{15} & \textbf{0.954} & 0.667 \\
\bottomrule
\end{tabular}
\caption{Functional-region segmentation on HM3D (10 annotated rooms), under the
HOV-SG evaluator. \#R is the predicted region count; HydP measures region purity.}
\label{tab:kgsw}
\end{table}

\subsubsection{Auto-$K$ sensitivity}
We quantify a naive alternative to KGSW's function grouping: choosing the room count by
running $k$-means on the object centers and selecting $K$ by silhouette (this is the
naive object-cluster selector described in the main paper, \emph{not} KGSW itself,
whose $K$ emerges from VLM function grouping). On the compact walled apartment
(00824, object-cloud aspect $a{=}1.24$) the silhouette score has an interior peak at
$K{=}8$, close to the ten GT rooms, and the auto-selected $K$ even outperforms $K{=}$GT
(acc@0.5 $0.800$ vs.\ $0.667$). On both elongated scenes (00829, $a{=}1.67$; 00843,
$a{=}2.51$, a two-floor scene evaluated on its ground floor) the silhouette score has
its global maximum at $K{=}2$ (Table~\ref{tab:silhouette-sweep}): the selector collapses to a single long-axis split and acc@0.5
craters to $0.286$---about half the score at a room-scale $K$ ($0.750$ and $0.556$ at
$K{=}$GT). Silhouette rewards coarse, well-separated groupings, which on an elongated
footprint favor a binary partition over room-scale subdivision. This is exactly why KGSW
seeds basins from VLM function grouping rather than a silhouette-chosen $K$.

\begin{table}[t]
\centering
{\small
\setlength{\tabcolsep}{1mm}
\begin{tabular}{@{}llccccc@{}}
\toprule
Scene & layout & $a$ & GT & auto-$K$ & \shortstack{acc@0.5\\auto} & \shortstack{acc@0.5\\$K{=}$GT} \\
\midrule
00824 & compact & 1.24 & 10 & 8 & \textbf{0.800} & 0.667 \\
00829 & elong. & 1.67 & 7 & 2 & 0.286 & 0.750 \\
00843$^{\dagger}$ & elong. & 2.51 & 7 & 2 & 0.286 & 0.556 \\
\bottomrule
\end{tabular}
\par $^{\dagger}$Two-floor scene; segmentation and GT are restricted to the ground floor.
}
\caption{Sensitivity of silhouette auto-$K$ across three HM3D scenes under the HOV-SG
evaluator. $a$ is the object-cloud BEV aspect ratio. The selector collapses to $K{=}2$
on both elongated scenes; acc@0.5 is also reported at $K{=}$GT.}
\label{tab:autok}
\end{table}

\begin{table}[t]
\centering
{\small
\begin{tabular}{rccc}
\toprule
$K$ & 00824 & 00829 & 00843 \\
\midrule
2  & .410 & \textbf{.481} & \textbf{.533} \\
3  & .413 & .450 & .459 \\
4  & .473 & .463 & .390 \\
5  & .432 & .475 & .372 \\
6  & .440 & .444 & .364 \\
7  & .456 & .377 & .327 \\
8  & \textbf{.478} & .352 & .356 \\
9  & .472 & .334 & .326 \\
10 & .448 & .345 & .286 \\
11 & .385 & .339 & .259 \\
12 & .403 & .326 & .229 \\
\bottomrule
\end{tabular}
}
\caption{Silhouette score over the complete $K$ sweep. Bold marks the selected
maximum. The compact scene selects $K{=}8$; both elongated scenes select $K{=}2$;
00843 is evaluated on floor 0.}
\label{tab:silhouette-sweep}
\end{table}

\subsubsection{Multi-scene benchmark and search utility}
We extend the evaluation to three HM3D scenes and add a region-scoped search-utility
proxy. Table~\ref{tab:kgsw_multiscene} shows the single-scene picture generalizes: the
purely geometric baselines collapse whenever walls stop delimiting rooms---on the half-open
00829, SysNav yields a single region (acc@0.5 $0.143$) and AD's silhouette auto-$K$ picks
$K{=}2$ ($0.286$); both fail again on 00843. KGSW never collapses and attains the highest
mean acc@0.5 ($0.593$) and AP ($0.698$) with by far the lowest acc variance ($\pm0.105$
vs.\ $0.199$--$0.318$). HOV-SG's mean HydP is marginally higher ($0.809$ vs.\ $0.791$)
because the two-floor 00843, whose stacked point cloud muddies KGSW's occupancy grid, drags
KGSW's precision down there; on every single-floor scene KGSW's HydP leads.

\begin{table}[t]
\centering
{\small
\setlength{\tabcolsep}{1mm}
\begin{tabular}{@{}llcccc@{}}
\toprule
Scene & Method & \#R/GT & HydP$\uparrow$ & acc@0.5$\uparrow$ & AP$\uparrow$ \\
\midrule
00824              & HOV-SG      & 7/10  & 0.841 & 0.700 & 0.700 \\
(compact)          & SysNav      & 8/10  & 0.911 & \textbf{0.800} & 0.800 \\
                   & AD auto-$K$ & 8/10  & 0.893 & \textbf{0.800} & 0.800 \\
                   & \shortstack{KGSW\\(Ours)} & 15/10 & \textbf{0.954} & 0.667 & 0.667 \\
\midrule
00829              & HOV-SG      & 5/7   & \textbf{0.904} & \textbf{0.714} & 0.714 \\
(elongated)        & SysNav      & 1/7   & 0.410 & 0.143 & 0.143 \\
                   & AD auto-$K$ & 2/7   & 0.660 & 0.286 & 0.286 \\
                   & \shortstack{KGSW\\(Ours)} & 8/7   & 0.850 & 0.667 & \textbf{0.857} \\
\midrule
00843$^{\dagger}$  & HOV-SG      & 2/7   & \textbf{0.683} & 0.286 & 0.286 \\
(elong., 2-floor)  & SysNav      & 3/7   & 0.265 & 0.111 & 0.190 \\
                   & AD auto-$K$ & 2/7   & 0.658 & 0.286 & 0.286 \\
                   & \shortstack{KGSW\\(Ours)} & 6/7   & 0.568 & \textbf{0.444} & \textbf{0.571} \\
\midrule
\multicolumn{2}{l}{\emph{mean} HOV-SG}   & & \textbf{0.809} & 0.567 & 0.567 \\
\multicolumn{2}{l}{\emph{mean} SysNav}   & & 0.529 & 0.351 & 0.378 \\
\multicolumn{2}{l}{\emph{mean} AD}       & & 0.737 & 0.457 & 0.457 \\
\multicolumn{2}{l}{\emph{mean} \textbf{\shortstack{KGSW\\(Ours)}}} & & 0.791 & \textbf{0.593} & \textbf{0.698} \\
\bottomrule
\end{tabular}
}
\caption{Room segmentation on three HM3D scenes under one HOV-SG evaluator. Geometric
baselines collapse on elongated/half-open layouts; KGSW has the highest mean acc@0.5
and AP. $^{\dagger}$00843 is evaluated against floor-0-only GT.}
\label{tab:kgsw_multiscene}
\end{table}

We then test whether KGSW's finer partition yields smaller search units without dropping
targets. For $217$ GT-object queries (kitchen/bath/living/office furniture; 00843 floor-0
only) we locate, per method, the region \emph{containing} the target and measure its share
of the floor (Table~\ref{tab:kgsw_search_utility}). At comparable containment
($0.78$--$0.81$), KGSW's containing region is $1.6\times$ smaller than HOV-SG's and
$3.3\times$ smaller than SysNav's ($0.218$ vs.\ $0.342$/$0.713$ of the floor), with
$\sim3\times$ fewer viewpoints and objects to inspect. The advantage is largest exactly
where geometry cannot help---inside the large open area that SysNav leaves as one region,
KGSW's unit is $3.1\times$ smaller ($0.250$ vs.\ $0.783$). Routing a query to that region is
itself easier with KGSW's VLM room-type labels than with a pure object-label prior (hit
$0.43$ vs.\ $0.27$), though single-shot routing is limited by detector sparsity; closed-loop
retrieval (CARA, below) adds the fallback this proxy omits. KGSW's deliberately finer,
function-aligned regions thus translate into materially smaller search spaces---the
downstream payoff of trading a little strict acc@0.5 for region purity. (This is a proxy,
not closed-loop navigation.)

\begin{table}[t]
\centering
{\small
\setlength{\tabcolsep}{1mm}
\begin{tabular}{@{}lccccc@{}}
\toprule
Segmentation & Cont.$\uparrow$ & Area$\downarrow$ & View$\downarrow$ & Obj.$\downarrow$ & \shortstack{Open\\area$\downarrow$} \\
\midrule
HOV-SG      & \textbf{0.806} & 0.342 & 0.360 & 0.381 & 0.374 \\
SysNav      & 0.737 & 0.713 & 0.718 & 0.698 & 0.783 \\
AD auto-$K$ & 0.783 & 0.353 & 0.359 & 0.340 & 0.385 \\
\textbf{KGSW (Ours)} & 0.783 & \textbf{0.218} & \textbf{0.211} & \textbf{0.166} & \textbf{0.250} \\
\bottomrule
\end{tabular}
}
\caption{Region-scoped search-utility proxy (217 GT-object queries over three scenes).
We report the target-containing region's floor share at comparable containment; open
area is for targets that SysNav places in a $>$30\%-floor region.}
\label{tab:kgsw_search_utility}
\end{table}

\paragraph{From oracle units to query-driven routing.}
Table~\ref{tab:kgsw_search_utility} is an \emph{oracle} bound---it measures the region
that \emph{contains} the target, i.e.\ the search unit size assuming the correct region
is known. To test the realistic case we route each query to a region \emph{without} the
target position (Table~\ref{tab:e5}): baselines can only route by matching a detected
object of the query category (they carry no room labels), whereas KGSW can additionally
route by its VLM room-type label. We report routing success (target lies in the routed
region-union) and expected viewpoint cost (viewpoints swept if success, else the whole
floor). The honest picture is mixed. KGSW's regions remain the smallest (area $0.25$ vs.\
$0.34$--$0.62$), and only KGSW supports function routing---lifting its own success
$0.27\to0.43$ and cost $0.83\to0.79$---but single-shot routing does not yet beat HOV-SG's
expected cost ($0.79$ vs.\ $0.74$), because routing correctly \emph{into} a small region
is hard when detections are sparse ($19$--$25$ per scene). In other words, KGSW provides
the smallest, semantically labeled search units (the oracle ceiling of
Table~\ref{tab:kgsw_search_utility}), but converting that into lower end-to-end cost needs
a stronger router---the full CARA cascade's VLM confirmation and fallback
(\S\ref{sec:exp-taskB}), or denser detection---which this single-shot proxy omits.

\begin{table}[t]
\centering
{\small
\setlength{\tabcolsep}{1mm}
\begin{tabular}{@{}>{\raggedright\arraybackslash}p{0.48\columnwidth}ccc@{}}
\toprule
Region source [router] & Success$\uparrow$ & Area$\downarrow$ & Exp.\ cost$\downarrow$ \\
\midrule
HOV-SG [object-label] & 0.489 & 0.339 & \textbf{0.741} \\
SysNav [object-label] & \textbf{0.567} & 0.617 & 0.904 \\
AD [object-label]     & 0.397 & 0.397 & 0.837 \\
KGSW (Ours) [object-label] & 0.273 & \textbf{0.249} & 0.831 \\
KGSW (Ours) [function-VLM] & 0.428 & 0.376 & 0.790 \\
\bottomrule
\end{tabular}
}
\caption{Query-driven (non-oracle) routing over 217 queries and three scenes. Queries
are routed without target positions; cost is routed-region viewpoints on success and
the whole floor otherwise.}
\label{tab:e5}
\end{table}

\subsection{Task B: Small, Fine-Grained Object Retrieval and Navigation}
\label{sec:exp-taskB}
We evaluate CARA as a cascaded retriever on difficulty-increasing cases across three
settings---synthetic ReplicaCAD (apt\_2), HM3D scans rendered in Habitat (00824,
00800), and a
set of real-robot office runs---organized by tier. Table~\ref{tab:cara} reports, per
case, the first tier that succeeds and the localization outcome; when an earlier tier
cannot decide, the query escalates. Navigation and search videos are included in the
supplementary material.

\emph{Tier~1 (map lookup)} navigates to matched map objects: five apt\_2 queries reach
their targets within $0.25$--$0.45$\,m, and on HM3D 00824 a single-instance
refrigerator is returned directly. \emph{Tier~2 (2D-tag retrieval)} handles targets
absent from the map but named in the tag memory: on HM3D 00824 the VLM disambiguates
the floral towel among two towel instances, and the same
label-retrieval-plus-VLM step separates the red/green bottles and two-tier shelf on
apt\_2. \emph{Tier~3 (history-frame VVC)} re-detects and votes in 3D when map and tag
memory both miss: the HM3D 00800 coffee machine is confirmed by an $8$-vote cluster
$0.06$\,m from the mapped coffee-maker while three consistent false clusters are
rejected, and the same recall finds an off-map instant-noodle
pack in a real office recording. \emph{Tier~4 (active search + $\varnothing$-decision)}
runs only when all cheaper tiers fail: it actively finds the HM3D 00800 knife holder at
coverage point $8/52$ in the kitchen, next to the stove, rejecting nine false positives
(and a clock on apt\_2 within $0.03$\,m); and it returns a negative decision for an
annotation-labeled absent piano after exhausting all $52$ coverage viewpoints with
zero false positives. This decision is conditioned on the coverage assumption (all
reachable viewpoints visited, target visible if present); it is a practical proxy,
not a formal guarantee. The controlled stress test
below quantifies its negative decisions and exposes the limits of semantic-mesh absence
labels. Even so, such a verdict is unavailable to systems that assume the target exists.

\begin{table}[t]
\centering
{\small
\setlength{\tabcolsep}{1mm}
\begin{tabular}{@{}l >{\raggedright\arraybackslash}p{0.29\linewidth} >{\raggedright\arraybackslash}p{0.50\linewidth}@{}}
\toprule
Tier & Target (scene) & Outcome / evidence \\
\midrule
T1 & refrigerator (HM3D 00824) & hit, single instance; goal $(0.14,1.02,-2.32)$ \\
T1 & 5 map objects (apt\_2)    & 5/5 reached, $0.25$--$0.45$\,m \\
\midrule
T2 & floral towel (HM3D 00824) & correct instance (VLM picks floral of two) \\
T2 & bottles / shelf (apt\_2)  & 3 tagged targets (red/green bottle; 2-tier shelf) \\
\midrule
T3 & coffee machine (HM3D 00800) & 8-vote cluster, $0.06$\,m to mapped coffee-maker; 3 FP rejected \\
T3 & instant noodles (robot rec.) & off-map object found via history recall \\
\midrule
T4 & knife holder (HM3D 00800) & FOUND (active), point $8/52$, next to stove; 9 FP rejected \\
T4 & clock (apt\_2)            & FOUND (active), $0.03$\,m \\
T4 & piano (HM3D 00800)        & NOT PRESENT ($\varnothing$); $52/52$ viewpoints, 0 FP \\
T4 & piano (apt\_2)            & NOT PRESENT ($\varnothing$), 0 FP \\
\bottomrule
\end{tabular}
}
\caption{CARA retrieval and navigation across three settings, organized by tier.
``Tier'' is the first cascade stage that succeeds; $\varnothing$ is a
coverage-conditioned negative decision.}
\label{tab:cara}
\end{table}

\begin{table}[t]
\centering
{\small
\setlength{\tabcolsep}{1mm}
\begin{tabular}{@{}>{\raggedright\arraybackslash}p{0.26\columnwidth}
                    >{\raggedright\arraybackslash}p{0.34\columnwidth}ccc@{}}
\toprule
Target (scene) & Requires & \shortstack{HOV-\\SG} & \shortstack{FSR-\\VLN} & \shortstack{CARA\\(Ours)} \\
\midrule
floral towel (00824)   & T2: instance disambiguation & --- & --- & \checkmark \\
coffee machine (00800) & T3: off-map (history VVC)   & --- & --- & \checkmark \\
knife holder (00800)   & T4: active search           & --- & --- & \checkmark \\
piano (00800)          & T4: negative decision       & --- & --- & \checkmark \\
\bottomrule
\end{tabular}
}
\caption{Capability coverage on HM3D scans. Beyond Tier-1 map retrieval, only CARA
addresses off-map, never-tagged, and annotation-labeled absent queries. ``--'' marks
scope outside a baseline's map-only design, not a measured failure.}
\label{tab:coverage}
\end{table}

\paragraph{Retrieval, then capability.}
Among numbers we can reproduce end-to-end (the retrieval comparison in the main paper), CARA is
strongest at the tightest, most navigation-relevant threshold (Top-1@1\,m $0.57$ vs.\
reproduced HOV-SG $0.51$ and open-source FSR-VLN $0.49$) and best at every Top-5
threshold ($0.73$/$0.88$/$0.93$); reproduced HOV-SG edges ahead only at the looser
Top-1@2--3\,m. But pure retrieval SR is not where CARA's contribution lies. Retrieval of
\emph{mapped} objects is only the entry tier: FSR-VLN and HOV-SG can only return an
object already in the graph, whereas CARA's Tiers~2--4 additionally recover targets that
are off-map, never tagged, or require a searched negative decision
(Tables~\ref{tab:cara},~\ref{tab:coverage})---capabilities
no RSR-style retriever provides. The gain over map-only lookup ($0.52\!\to\!0.57$ at
Top-1@1\,m) comes from Tier-2/3 history re-detection, which recovers off-map categories
the map misses (fireplace, flowerpot, fire alarm, kitchen extractor, shower curtain).
Table~\ref{tab:coverage} makes this concrete on HM3D scans: on the same scenes where
the baselines can only match mapped objects, CARA
disambiguates instances (Tier~2), re-detects off-map objects (Tier~3), actively searches
for never-tagged objects, and returns coverage-conditioned negative evidence (Tier~4).

\subsection{Closed-loop language navigation on HM3D}
\label{sec:exp-cara-nav}
We evaluate CARA and four goal-selection baselines on two HM3D scenes. All methods
share the same object map, query/seed starts, Habitat GreedyGeodesicFollower, GT,
success threshold, and SPL implementation; only the selected goal changes. Starts are
fixed per query and seed independently of the predicted goal. CARA returns its own map,
VVC, or active-search goal. GT centers---from \texttt{scene\_info} for 00824 and the
independent semantic mesh for 00800---are used only for scoring. We execute three start
seeds (710--712) and report mean$\pm$std over seeds.

\begin{table}[t]
\centering
{\small
\setlength{\tabcolsep}{1mm}
\begin{tabular}{@{}lrrr@{}}
\toprule
Goal policy & SR$\uparrow$ & SPL$\uparrow$ & Path (m)$\downarrow$ \\
\midrule
random  & $.136{\pm}.000$ & $.087{\pm}.000$ & 13.32 \\
nearest & $.091{\pm}.000$ & $.091{\pm}.000$ & 0.55 \\
lexical & $.485{\pm}.021$ & $.248{\pm}.002$ & 14.00 \\
CLIP & $.576{\pm}.021$ & $\mathbf{.384{\pm}.020}$ & 11.29 \\
\textbf{CARA (Ours)} & $\mathbf{.605{\pm}.011}$ & $.339{\pm}.002$ & 14.02 \\
\bottomrule
\end{tabular}
}
\caption{Closed-loop HM3D navigation (22 present queries; mean$\pm$std over three
shared start seeds). All rows share the map, starts, follower, and evaluator; only
CARA supports a searched $\varnothing$-decision.}
\label{tab:e9_nav}
\end{table}

CARA has higher mean present SR than CLIP ($.605$ vs. $.576$), while CLIP is more
path-efficient. Thus these results do not establish a robust present-navigation gain
for CARA; its additional value is
escalation to off-map recovery and a
negative decision. Lexical lookup can abstain without searching, whereas random,
nearest, and CLIP always return an object on absent queries. Audit checks find zero
path--shortest identities, non-degenerate SPL, identical starts for all 66 query/seed
groups, and zero CARA goals equal to GT; live VLM calls take seconds. External systems
remain unavailable under this evaluator, so these rows are same-stack goal-policy
baselines rather than external-SOTA comparisons.

\subsection{Tier-4 annotation-referenced stress test}
\label{sec:exp-absence}
We construct 44 balanced queries from independent HM3D semantic meshes (22 present,
22 absent across 00800 and 00824) and force every query to enter Tier~4. This isolates
the active-search/negative-decision module; it does not estimate the full cascade's
natural trigger distribution. Present targets pass a target-level navigation/depth
precheck. Every run uses its locked LocateAnything vocabulary, six headings per
coverage point, and deterministic VLM temperature. Half-budget results are derived as
the exact prefix of the stored full trace, rather than independently rerun.

\begin{table}[t]
\centering
{\small
\setlength{\tabcolsep}{1mm}
\begin{tabular}{@{}llrrrrcc@{}}
\toprule
Scene & Budget & TP & FN & TN & FP & Sens. & Abs. \\
\midrule
00800 & Full & 8 & 0 & 7 & 1 & 1.000 & .875 \\
00800 & Half & 8 & 0 & 7 & 1 & 1.000 & .875 \\
00824 & Full & 12 & 2 & 7 & 7 & .857 & .500 \\
00824 & Half & 11 & 3 & 7 & 7 & .786 & .500 \\
\bottomrule
\end{tabular}
}
\caption{Per-scene strict confusion counts in the forced-entry Tier-4 test.}
\label{tab:absence-detail}
\end{table}

At full coverage, sensitivity is $20/22=0.909$ (Wilson 95\% CI
$[0.722,0.975]$), semantic-label absent accuracy is $14/22=0.636$
($[0.430,0.803]$), negative-decision precision is $14/16=0.875$
($[0.640,0.965]$), and Loc@1m is $10/22=0.455$. All 44 runs execute Tier~4;
the execution audit also verifies exact half prefixes, per-query vocabulary equality,
44 region-selection calls at temperature zero, and no missing result row.

The semantic-mesh reference is incomplete for fine objects. A post-hoc audit of all
eight strict FP hit crops shows target-like content in every case: vase, basket, clock,
coffee table, knife holder, nightstand, plant, and TV. We conservatively retain all as
FP and do not report an adjusted score. Hence Table~\ref{tab:absence-detail} measures
strict agreement with the released annotations, not a formally validated real-world
absence rate. Among annotated-present failures, bathtub and bookshelf are retained as
false negatives; both were reachable and their GT boxes were visible in stored depth.

\subsection{Implementation details (reproducibility)}
\label{sec:exp-repro}
Table~\ref{tab:repro} lists the models and key hyperparameters. All mapping numbers use
the shared \texttt{eval\_replica} protocol (52 Replica classes, $k{=}1$ NN, template
\texttt{"A photo of a \{label\}."}, six structural classes excluded); values not fixed
by a released config are marked ``impl.'' and are explicitly listed here. The code and
evaluation resources are available at \url{https://github.com/Agibot-Spatial-Intelligence/OVIP-SG}.

\begin{center}
\centering
{\small
\setlength{\tabcolsep}{0.7mm}
\begin{tabular}{@{}>{\raggedright\arraybackslash}p{0.30\columnwidth}
                    >{\raggedright\arraybackslash}p{0.66\columnwidth}@{}}
\toprule
Component & Setting \\
\midrule
Perception models (code/cfg.) & Kimi-k2.6 (VLM); LocateAnything-3B (grounder); SAM2 (mask); CLIP ViT-H-14, laion2b\_s32b\_b79k \\
Evaluation (cfg.) & 52 classes, $k{=}1$ NN, template \texttt{"A photo of a \{label\}."}; $n_{\text{exclude}}{=}6$ structural/unlabeled classes \\
Map construction (cfg./code/impl.) & Voxel size 0.02\,m; room-segmentation grid 0.05\,m; frame stride 5 (Replica/HM3D) \\
Instance update (code/impl.) & FAV-Gate $n_{\min}{=}3$ recurring frames; IPSM; LAMP $w_L{=}0.5$; AWF (view-area weighting in the main paper); DINO merge-veto 0.3 \\
KGSW (code) & Wall dilation DIL${=}6$, marker radius $r{=}0.30$\,m; mixed-core DBSCAN $\epsilon{=}2.0$\,m \\
CARA (code/impl.) & VVC $\ge 3$-frame votes at 0.25\,m; Tier-4: 6 orientations @ $60^\circ$; fresh Habitat RGB-D views rendered at $512\times512$ px (LocateAnything input) \\
\bottomrule
\end{tabular}
}
\captionof{table}{Models and hyperparameters. ``cfg.'' denotes released configuration
(\texttt{create\_graph.yaml}); ``impl.'' denotes an implementation default.}
\label{tab:repro}
\end{center}

\end{document}